# PARTITIONING RELATIONAL MATRICES OF SIMILARITIES OR DISSIMILARITIES USING THE VALUE OF INFORMATION

Isaac J. Sledge,[1] *Student Member, IEEE*    José C. Príncipe,[1,2] *Life Fellow, IEEE*

[1] Department of Electrical and Computer Engineering, University of Florida
[2] Department of Biomedical Engineering, University of Florida

**ABSTRACT**

In this paper, we provide an approach to clustering relational matrices whose entries correspond to either similarities or dissimilarities between objects. Our approach is based on the value of information, a parameterized, information-theoretic criterion that measures the change in costs associated with changes in information. Optimizing the value of information yields a deterministic annealing style of clustering with many benefits. For instance, investigators avoid needing to a priori specify the number of clusters, as the partitions naturally undergo phase changes, during the annealing process, whereby the number of clusters changes in a data-driven fashion. The global-best partition can also often be identified.

## 1. INTRODUCTION

The clustering of vector-based data is a critical problem, as it is encountered in many applications that involve analysis with little to no prior knowledge about the data [1]. The clustering of similarity- and dissimilarity-based relational data is also important [2]. A given representation of objects may not be readily defined in terms of features yet it can be characterized by the relationships between the objects [3–5]. This type of representation is common in many fields, including bioinformatics, computer vision, and psychology.

Regardless of the data representation, clustering is often formulated by defining a cost function to be minimized. Traditional approaches for optimizing these cost functions rely on coordinate descent to produce partitions of the data. Such approaches tend to converge only to sub-optimal solutions, as many of the cost functions are non-convex and hence contain many local minima.

One way to circumvent becoming trapped in local minima during clustering is to employ simulated annealing [6]. Simulated annealing works by generating a sequence of random partitions. The decision to accept a given partitions depends on the probability of the resulting configuration. The cost is therefore not always always minimized in a monotonic fashion. The process may iteratively jump from one local-best solution to a region with a worse one.

In the limit, simulated annealing will eventually reach a global minimizer. It requires a sufficiently slow cooling of parameters that influence the sequence generation for this to occur, though. This slow rate can be non-conducive for some applications.

Another option for avoiding local minima during clustering is to rely on deterministic annealing [7]. Deterministic annealing bears some resemblance to simulated annealing. It inherits some positive attributes of simulated annealing. It is guaranteed to reach the global minimum, for instance. This occurs despite replacing the random walk approach to generating partitions with an expectation. It also comes with other benefits. For example, while an adjustment of parameter values is necessary during the optimization process, the cooling rate can be much quicker than in simulated annealing. This makes it attractive for many real-world applications.

Due to its beneficial properties, deterministic annealing has been applied to the problem of clustering vector-based data. Several information-theoretic clustering algorithms are given by Rose and his colleagues [8, 9]. In each of these frameworks, they obtained a parameterized, Shannon-entropy-based [10] free-energy expression that describes the quality of a particular partition. They have shown that the free energy is minimized by the most probable set of cluster representatives at a given parameter value. At high parameter values, there is only one local minimum, which is global by default. This minimum corresponds to a crisp partition where each vector-based data point belongs to a single cluster. As the parameter value is lowered, more clusters emerge and the cluster memberships become increasingly fuzzy. A hierarchy of partitions, with decreasing average costs, is obtained as the process undergoes a series of phase transitions. The annealing process tracks the global minimum across each phase change.

While deterministic annealing has proved useful for clustering vectorial data, there has yet to be a formulation of it for clustering relational representations. In this paper, we provide an information-theoretic formulation [11] for such data types, which is the value of information due to Stratonovich [12].

The value of information [13, 14] is a type of free-energy criterion that describes the largest reduction in costs associated with a given amount of information. In the context of clustering, the number of groups is implicitly dictated by the information amount. High amounts of information lead to a small number of clusters with many elements. A potentially good qualitative partitioning of the relationships is often observed in such cases. Lower amounts of information yield larger number of clusters with fewer elements per cluster. The partitions can be qualitatively poor, as clusters are unnecessarily split. Determining the 'right' amount of information is therefore crucial.

When optimized, the value of information yields a deterministic annealing process for updating the cluster memberships. A hierarchy of partitions, corresponding to differing amounts of information, is produced through the annealing process. Partitions from this hierarchy that quantize the data well can be automatically identified through analysis of a rate-distortion-like curve. This allows investigators to sidestep needing to a priori specify the number of clusters. The cluster count, and hence the 'right' amount of information, is determined in a data-driven fashion. Our approach does not require manually setting any parameters, which is a novelty of our method compared to existing, vector-data-based deterministic annealing clustering schemes.

Unlike the approaches given by Rose et al. [8, 9], the value of information relies on Shannon mutual information [10], not Shannon entropy. The resulting partition update equations therefore contain extra terms that account for the cluster population statistics. These additional terms lead to more qualitatively appealing partitions of the data. They also avoid producing coincident clusters as the annealing process undergoes phase changes. The approaches of Rose et al. sometimes yield coincident clusters; this would also occur if we extended their work to the relational-data case, which we show.

## 2. METHODOLOGY

Our approach for relational clustering can be described as follows. Given a relational-matrix-based representation of a weighted graph, we seek to partition it to produce a reduced-size graph, which we refer to as an accumulation matrix. The vertices in the accumulation matrix are are analogous to cluster prototypes in the vector-data case. There is a one-to-many mapping of a vertex from the accumulation matrix to the vertices of the relational matrix. The edges of the accumulation matrix

This work was funded by ONR grant N00014-15-1-2013. The first author was additionally funded by a University of Florida Research Fellowship, a Robert C. Pittman Fellowship, and an ASEE Naval Research Enterprise Fellowship.

codify the relative dissimilarity between pairs of prototypes.

There are many possible accumulation matrices that can be formed for a given relational matrix. We would like to find one that minimizes some distortion measure. But, due to the different sizes of the relational and accumulation matrices, defining a distortion is difficult. We therefore specify how to construct a so-called composite matrix, that sidesteps this difficulty, and define an appropriate distortion between it and the original relational matrix. We provide an objective function for finding the optimal accumulation matrix from both the composite matrix and the relational matrix.

We also encounter another issue: uncovering the optimal accumulation matrix is not trivial due to the binary-valuedness of the one-to-many mappings. We relax the binary assumption and offer an alternate objective function, which is based on the value of information. Optimization of the value of information yields a deterministic annealing process for finding part of the accumulation matrix. In the limit, the solution of deterministic annealing approaches the global solution of the original objective function. A hierarchy of possible partitions, each with a different number of clusters, are produced as intermediate solutions.

## 2.1 Preliminaries

In what follows, we rely on the concept of a relational matrix. A relational matrix is an adjacency-based representation of a directed graph.

**Definition 1.1.** A relational matrix $R$ is a matrix $\mathbb{R}_+^{n \times n}$ given by $(V_\pi, E_\pi, \Pi)$, where $V_\pi$ is a set of $n$ vertices, $E_\pi$ is a set of edge connections between all pairs of vertices, and $[\Pi]_{i,j} = \pi_{i,j}$, are positive, symmetric, reflexive weights assigned to the graph edges. The subscripts on the vertices and edges represent the dependence on the particular weight matrix.

For any relational matrix, we define the notion of an outgoing vector $\pi_{i,1:n} = [\pi_{i,1}, \ldots, \pi_{i,n}]$ for the $i$th vertex by its weights on the outgoing edges. We assume that $\pi_{i,j} = 0$ in the outgoing vector if and only if there is no directed edge from the $i$th vertex to the $j$th vertex.

The outgoing vectors provide basis of comparison between vertex pairs. If two relational matrices, $R_\pi$ and $R_\varphi$ are of the same size, then this comparison can be performed on $\pi_{i,1:n}$ and $\pi_{j,1:n}$ according to a measure $g : \mathbb{R}_+^n \times \mathbb{R}_+^n \to \mathbb{R}_+$. If, however, they are of different sizes, then a composite matrix must be formed so that the distortion between the two matrices can be assessed.

**Definition 1.2.** A partition function $\psi$ is a mapping between two index sets such that $\psi^{-1}(\mathbb{Z}_{1:m})$ is a partition of $\mathbb{Z}_{1:n}$. That is, $\psi^{-1}(j) \subset \mathbb{Z}_{1:n}, \psi^{-1}(j) \cap \psi^{-1}(k) = \emptyset$, for $j \neq k$, and $\psi^{-1}(1) \cup \ldots \cup \psi^{-1}(m) = \mathbb{Z}_{1:n}$.

It can be seen that a partition induces a binary accumulation matrix $[\Psi]_{i,j} = \psi_{i,j}$, where $\psi_{i,j} = 1$ if $i \in \psi^{-1}(j)$ and $\psi_{i,j} = 0$ if $i \notin \psi^{-1}(j)$. Therefore, $[\Psi]_{1:n,k} = \sum_{i \in \psi^{-1}(k)} e_i$, where $e_i$ is the $i$th unit vector.

**Definition 1.3.** Given relational matrices $(V_\pi, E_\pi, \Pi)$, with $n$ vertices, and $(V_\varphi, E_\varphi, \Phi)$, with $m$ vertices, $R_\vartheta$ is the composite relational matrix $(V_\vartheta, E_\vartheta, \Theta)$, which satisfies the conditions

(i) The vertex set $V_\vartheta = V_\pi \cup V_\varphi$ is the union of all vertices in $R_\pi$ and $R_\varphi$. For simplicity, the composite vertex set is indexed such that the first $m$ nodes are from $R_\varphi$ and the remaining $n$ nodes are from $R_\pi$.

(ii) The edges in $R_\vartheta$ are one-to-many mappings from the vertices in $R_\varphi$ to the vertices in $R_\pi$. Each vertex in $R_\varphi$ represent groups of vertices from $R_\pi$. Although $R_\vartheta$ has $m+n$ vertices, we can represent its weighting matrix by $\Theta = [\vartheta_{1,1:m}^\top, \vartheta_{2,1:m}^\top, \ldots, \vartheta_{m,1:m}^\top]^\top$. The outgoing vectors $\vartheta_{i,1:n}$ are of the same direction as $\pi_{i,1:n}$.

(iii) The partition function $\psi$ provides an accumulation relation between the edge weights of $R_\varphi$ and $R_\vartheta$, which is given by $\varphi_{j,k} = \sum_{i \in \psi^{-1}(k)} \vartheta_{j,k}, \forall j, k$.

Note that, for our application, the two relational matrices will almost always be of different sizes. The first matrix, $R_\pi$, will be the matrix specified by an investigator. The accumulation matrix, $R_\varphi$, will essentially be a partitioning of $R_\pi$. The objects of $R_\varphi$ correspond to relational cluster prototypes; the weights of $R_\varphi$ correspond to prototype-prototype distances. Each object in $R_\pi$ will map to some prototype in $R_\varphi$.

We can now assess the distortion of any $R_\pi$ and $R_\varphi$. First, we define the weighted distance between corresponding outgoing vectors of $R_\pi$ and the composite matrix $R_\vartheta$ assigned by the partition $\psi$,

$$\sum_{i=1}^n p(i) g(\pi_{i,1:n}, \vartheta_{\psi(i),1:n}).$$

Here, $p(i)$ are a set of weights, which can be viewed as probabilities. We then use this weighted distance to define the quantization distortion between $R_\pi$ and $R_\varphi$, which is

$$q(R_\pi, R_\varphi) = \min_{R_\vartheta \in \mathbb{R}_+^{m \times n}} \left( \sum_{i=1}^n p(i) g(\pi_{i,1:n}, \vartheta_{\psi(i),1:n}) \,\middle|\, R_\vartheta \in R_{\pi\varphi} \right)$$

where $R_{\pi\varphi}$ is the set of all composite matrices for $R_\pi$ and $R_\varphi$. This objective function is over all possible sets of binary partitions.

Suppose that we have a relational matrix $R_\pi \in \mathbb{R}_+^{n \times n}$. We would like to find another relational matrix $R_\varphi \in \mathbb{R}_+^{m \times m}$ that provides a coarse representation of $R_\pi$. $R_\varphi$ is referred to as an accumulated relational matrix.

**Definition 1.4.** Suppose that we have a relational matrix $R_\pi = (V_\pi, E_\pi, \Pi)$ with $n$ vertices, where the weight matrix is given by $\Pi = [\pi_{1,1:n}^\top, \pi_{2,1:n}^\top, \ldots, \pi_{n,1:n}^\top]^\top$. An accumulation relational matrix is given by another relational matrix $R_\varphi = (V_\varphi, E_\varphi, \Phi)$ that has $m$ vertices, with a weight matrix $\Phi = [\varphi_{1,1:m}^\top, \varphi_{2,1:m}^\top, \ldots, \varphi_{m,1:m}^\top]^\top, m \leq n$. An accumulation relational matrix satisfies

$$\arg\min_{\Psi \in \mathbb{R}_+^{n \times m}, R_\varphi \in \mathbb{R}_+^{m \times m}} \left( q(R_\pi, R_\varphi) \,\middle|\, [\Psi]_{1:n,k} = \sum_{i \in \psi^{-1}(k)} e_i \right),$$

for the positive distortion measure $q : \mathbb{R}_+^{n \times n} \times \mathbb{R}_+^{m \times m} \to \mathbb{R}_+$ given above.

It is immediate that at least one minimizer for this function exists, since the number of possible binary partitions is finite. Due to the binary nature of the partitions, finding an accumulation relational matrix has an NP-hard computational complexity.

## 2.2 Partitioning Relational Matrices

We hence seek approximately optimal accumulation relational matrices that are more computationally tractable to produce. To do this, we decompose the optimization problem in definition 1.4, which is outlined by the following definition.

**Definition 1.5.** Suppose that we have a relational matrix $R_\pi = (V_\pi, E_\pi, \Pi)$ with $n$ vertices, where the weight matrix is given by $\Pi = [\pi_{1,1:n}^\top, \pi_{2,1:n}^\top, \ldots, \pi_{n,1:n}^\top]^\top$. Suppose that we also have a composite relational matrix $R_\vartheta = (V_\vartheta, E_\vartheta, \Theta)$ with $n+m$ vertices, where the weight matrix is given by $\Theta = [\vartheta_{1,1:m}^\top, \vartheta_{2,1:m}^\top, \ldots, \vartheta_{m,1:m}^\top]^\top$.

An accumulation relational matrix $R_\varphi = (V_\varphi, E_\varphi, \Phi)$ that has $m$ vertices, can be constructed in a two-step fashion:

(i) Vertex grouping: Solve the optimization problem given in (1) for the positive distortion measure $q$ given above. This has the effect of partitioning the $n$ vertices of the relational matrix $R_\pi$ into $m$ groups. To each group, a representative super-vertex is ascribed such that the average pairwise distance between a vertex and a super-vertex is minimized.

(ii) Edge aggregation: Obtain $R_\varphi$ from the following expression: $\varphi_{j,k} = \sum_{i \in \psi^{-1}(k)} \vartheta_{j,k}$ using the optimal weights $\vartheta_{i,j}$ and partition matrix $\Psi$ from step (i).

To address the vertex grouping problem in the first step, we utilize the value of information. The value of information is an information-theoretic criterion originally proposed by Stratonovich. We have previously shown how it can be used for addressing the exploration-exploitation problem in reinforcement learning [15–17]. We have also

$$\arg\min_{\Psi\in\mathbb{R}_+^{n\times m},\, R_\vartheta\in\mathbb{R}_+^{m\times n}} \left( \sum_{i=1}^n p(i) g(\pi_{i,1:n}, \vartheta_{\psi(i),1:n}) \,\Bigg|\, R_\vartheta \in R_{\pi\varphi},\, [\Psi]_{1:n,k} = \sum_{i\in\psi^{-1}(k)} e_i \right) \quad (1)$$

$$\min_{P\in\mathbb{R}_+^{m\times n}} \left( \min_j \left( \sum_{i=1}^n p(i) g(\pi_{i,1:n}, \varphi_{j,1:m}) \right) - \left( \sum_{i=1}^n \sum_{j=1}^m p(i)p(j|i) g(\pi_{i,1:n}, \varphi_{j,1:m}) \right) \,\Bigg|\, \varphi_{j,k} = \sum_{i=1}^n \vartheta_{i,k} p(j|k) \right) \quad (2)$$

demonstrated that using the value of information leads to a clustering of the state-action space according to the value function.

For the problem of clustering relational data, the value of information can be used to quantify the expected amount of decrease in the matrix-matrix distortion associated with changes in information. Information, in a clustering context, corresponds to how finely we are partitioning the data. Low amounts of information correspond to many clusters and fuzzy cluster memberships. High amounts of information correspond a single cluster and crisp cluster memberships. The choice of the 'right' amount of information for a given dataset, and hence the number of clusters, can be made automatically by processing a rate-distortion-like curve that relates the distortion of $R_\pi$ and $R_\varphi$ to information.

We utilize the value of information, given in (2), to induce a soft partitioning of the relational matrix. This leads to the notion of a soft accumulation matrix.

**Definition 1.6.** Suppose that we have a relational matrix $R_\pi = (V_\pi, E_\pi, \Pi)$ with $n$ vertices, where the weight matrix is given by $\Pi = [\pi_{1,1:n}^\top, \pi_{2,1:n}^\top, \ldots, \pi_{n,1:n}^\top]^\top$. A soft accumulation relational matrix is given by another relational matrix $R_\varphi = (V_\varphi, E_\varphi, \Phi)$ that has $m$ vertices, with a weight matrix $\Phi = [\varphi_{1,1:m}^\top, \varphi_{2,1:m}^\top, \ldots, \varphi_{m,1:m}^\top]^\top$, $m \leq n$. A soft accumulation relational matrix satisfies (2), where $[P]_{j,i} = p(j|i)$ is a set of non-negative association weights that take values over the unit interval. Such weights are subject to a Shannon mutual information constraint

$$\left( \sum_{i=1}^n \sum_{j=1}^n p(i)p(j|i)\log p(j|i) \right) - \left( \sum_{i=1}^n \sum_{j=1}^n p(i,j)\log p(j) \right) \leq r$$

where $r \geq 0$ is a user-specified parameter.

The problem of finding soft accumulation matrices is specified by the above constrained optimization problem. To effectively solve this problem, we form the Lagrangian and differentiate it. This provides a grouped coordinate descent procedure for specifying the non-negative association weights.

**Proposition 1.1.** For a given relational matrix $R_\pi$, the accumulation relational matrix $R_\varphi$ can be found from the non-negative association weights determined by the following alternating updates

$$p^{(k)}(j) \leftarrow \sum_{i=1}^n p(i)p(j|i),$$

$$p(j|i) \leftarrow p(j)e^{-\beta g(\pi_{i,1:n},\vartheta_{j,1:m})} \Big/ \sum_{j=1}^m p(j)e^{-\beta g(\pi_{i,1:n},\vartheta_{j,1:m})},$$

which are iterated until convergence. Here, $\beta \geq 0$ is a Lagrange multiplier that emerges from the Shannon mutual information constraint.

The variable $p(j)$ corresponds to the cluster population statistics.

Substituting the association weights from proposition 1.1 into the Lagrangian yields

$$F(R_\pi, R_\vartheta) = -\frac{1}{\beta} \sum_{i=1}^n p(i) \log \left( \sum_{j=1}^m e^{-\beta g(\pi_{i,1:n},\varphi_{j,1:m})} \right).$$

At each grouped-coordinate descent iteration, the Lagrange multiplier $\beta$ is fixed and a local minimum of the Lagrangian is found. That is, the representative outgoing vectors $\vartheta_{j,1:m}$ are computed using the following implicit equation

$$\nabla_{\vartheta_{j,1:m}} F(R_\pi, R_\vartheta) = \sum_{i=1}^n p(i)p(j|i) \nabla_{\vartheta_{j,1:m}} g(\pi_{i,1:n}, \vartheta_{j,1:m}) = 0.$$

This equation can be solved using gradient descent methods where the solutions from the previous iterations are used as the starting values for the current iteration. These computations are repeated as the multiplier $\beta$ is increased, leading to an annealing-like process. For small values of $\beta$, this procedure finds the global minimum of the Lagrangian. This minimum is tracked as $\beta$ is iteratively increased.

The effects of $\beta$ are as follows. As $\beta$ tends to zero, minimizing the Lagrangian is approximately same as minimizing the negative Shannon information. Shannon information is known to be convex and hence has a global minimizer. In this case, the weights are approximately uniform, $p(j|i) \approx m^{-1}\,\forall i,j$, so all outgoing vectors $\vartheta_{j,1:m}$ are coincident. There are hence many clusters, and every object in $R_\pi$ has the same fuzzy membership to each cluster.

As $\beta$ is increased, the soft accumulation matrix becomes more crisp. Smaller number of clusters are formed. Moreover, the annealing process exhibits a series of phase transitions where the outgoing vectors $\vartheta_{j,1:m}$ are insensitive to changes in $\beta$ except at critical values. The number of distinct outgoing vectors in the composite relational matrix increases at these critical values. When $\beta$ approaches infinity, the information constraint is essentially ignored. Thus, minimizing the Lagrangian is the same as minimizing the relational-matrix distortion $q$ between $R_\pi$ and $R_\varphi$. We therefore obtain an almost-crisp partition, $p(j|i) \approx 1\,\forall i,j$. We also begin to recover the relational-matrix distortion function over binary partitions. This crisp partition contains only a single cluster.

**Determining Number of Clusters.** Our approach to clustering relational matrices entails iterating over a range of $\beta$ values from small to large. Each value of $\beta$ between two critical values leads to accumulation matrices that define partitions with different number of clusters. This entire process yields a hierarchy of partitions.

Investigators are often interested in obtaining only a single partition, containing the 'right' number of clusters, that 'best' fits the observations. We hence consider an automated heuristic of choosing a parsimonious partition from the hierarchy that is produced.

Our heuristic is based on comparing the amount of information $r$ against the relational-matrix dissimilarity between $R_\pi$ and $R_\varphi$. Such a comparison leads to a rate-distortion like curve, which often contains a knee-like region for some moderate information amount. Our studies have shown that partitions around the knee region often qualitatively partition the data well with few to no unnecessary clusters. A good partition in this region can be easily detected via:

(i) Iterating over each point along this curve. For each point, we fit two linear functions that bisect it: one of which is a least-squares fit to the part of the curve that is to the left of the bisector and one that is fit to the part of the curve to the right.

(ii) Finding the point that leads to the lowest sum of least-squared errors for the two linear functions, which almost always corresponds to the knee. The partition corresponding to this amount of information $r$ (as quantified by $\beta$) is returned.

## 3. EXPERIMENTS

To assess our approach, we consider three relational datasets from real-world applications. We have previously analyzed these datasets in [18, 19], in the context of relational cluster validity.

(i) $R_{\text{GD-30}}$: This data was formed from a combination of cDNA microarray gene expressions and gene ontology similarities of 30 genes related to cell apoptosis in human lymphomas.

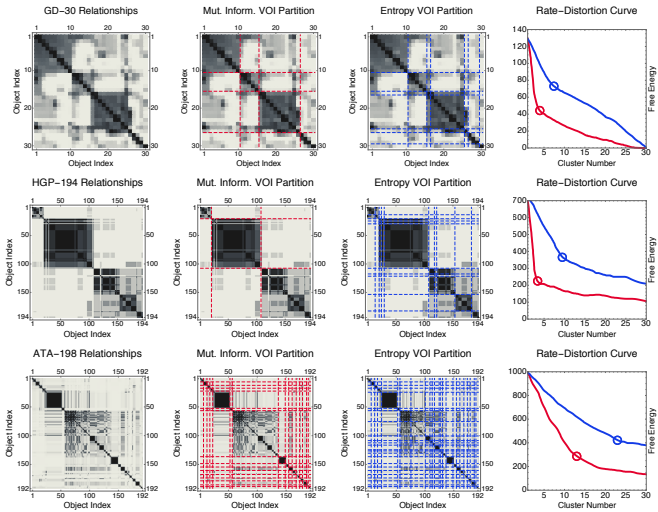

Figure 1: Deterministic annealing clustering results for three datasets: $R_{\text{GD-30}}$, $R_{\text{HGP-194}}$, and $R_{\text{ATA-198}}$. For each row, the first plot is the relational matrix of dissimilarities. Dark colors correspond to low dissimilarity, while lighter colors correspond to higher dissimilarity. The second and third plots are the crisp partitions returned for out method when using Shannon mutual information and Shannon entropy, respectively, to specify the constraint for the value of information (VOI) criterion. We converted the fuzzy partitions to crisp partitions for display purposes. The fourth plot provides a rate-distortion curve which compares the free-energy magnitude to the amount of information. We converted the information amount to the number of clusters to ease interpretations of these plots. Circles in these plots are provided to highlight the number of clusters chosen by our automated knee-detection heuristic.

(ii) $R_{\text{HGP-194}}$: This data was constructed by applying gene ontology similarity measures to 194 human gene products. The data are composed of three groups or 21, 86, and 87 gene products from the myotubularin, receptor precursor, and collagen alpha chain protein families, respectively.

(iii) $R_{\text{ATA-198}}$: This data was created from a combination of gene ontology similarity and microarray gene expressions on 198 genes from the *Arabidopsis Thaliana* plant. The plant was subject to a variety of stresses from insects and stress controls.

Note that the relationships for each of these datasets cannot be considered as pairwise distances between latent set of vectors. They were generated completely from similarity measures. We hence cannot expect to perform multi-dimensional scaling [20, 21], to be able to apply vector-data deterministic annealing clustering algorithms, without disrupting the cluster structure [22].

### 3.1 Results and Discussions

We applied our clustering approach to the above three relational datasets. Note that our approach has no parameters that must be manually set. Corresponding results are presented in figure 1.

For each dataset, we thresholded the fuzzy association weights and overlaid the resulting crisp partition on top of the relationships. We expect to see these partitions segment the dark, blocky structures along the main diagonal. These blocky structures correspond to compact, low-dissimilarity object groups [?]. Light values on the off-diagonal indicate that these groups are well separated. We re-ordered the dissimilarities according to the visual assessment of cluster tendency algorithm [23–25] to better highlight the latent data structure.

In figure 1, we also provide results for the case where Shannon entropy is used as a constraint for the free-energy criterion instead of Shannon mutual information. This is a direct extension of Rose's deterministic annealing method for the relational-data case.

**Clustering Results.** For $R_{\text{GD-30}}$ and $R_{\text{HGP-194}}$, the value of information with the Shannon mutual information constraint returned a fuzzy partitions with $c=4$ and $c=3$ clusters, respectively. These partitions are consistent with the dark blocks present along the main diagonal; compact, well-separated clusters are therefore being properly identified. As we explain in the online appendix[1], this partitioning of genes best aligns with their biological functionality.

$R_{\text{ATA-198}}$ was, comparatively, more challenging due to the sparse nature of the gene relationships. There were genes that naturally grouped into many small clusters that were compact and well-separated. A total of $c=12$ clusters were identified by our method. This partitioning aligns with the visual interpretation of cluster structure according to the re-ordered dissimilarity plot: compact, well-separated groups are properly segmented. Some of these clusters also have biological significance, as we explain in the online appendix.

The value of information with the Shannon entropy constraint returned fuzzy partitions with $c=9$, $c=12$, and $c=24$ clusters, respectively, for $R_{\text{GD-30}}$, $R_{\text{HGP-194}}$, and $R_{\text{ATA-198}}$. This approach had the tendency to over-segment the data. Objects that were slight outliers were frequently assigned to a singleton cluster, which led to a large number of 'unnecessary' clusters. Coincident partitions were also returned, which redundantly described the natural object groupings. Manually aggregating these coincident clusters led to similar object groupings as the mutual-information-constrained approach.

**Cluster Validity Results.** To quantitatively assess the goodness of our results, we applied twenty relational cluster validity indices [18] to the hierarchy of crisp and fuzzy partitions produced. These included the generalized Dunn's indices [26], modified Hubert's statistics [27], and the Xie-Beni index [28].

For $R_{\text{GD-30}}$, almost every index selected $c=4$ as the best estimate for the number of clusters. For $R_{\text{HGP-194}}$, a majority of these indices chose $c=3$ as the best cluster count estimate. These results agree with both the visual partitioning of the data and the corresponding knees of the rate-distortion curves. We can hence conclude that our approach is identifying the cluster structure well for these two datasets.

Some of the validity indices for $R_{\text{GD-30}}$ favored partitions with $c=6$ or 7 clusters. Such partitions separated each of the genes in the bottom, right corner of the relational matrix into singleton clusters. Likewise, for $R_{\text{HGP-194}}$, some indices selected partitions with $c=4$ or $c=5$. These partitions identified the cluster sub-structure for the bottom-right block. The conflicting findings for both datasets are not necessarily incorrect, as there is biological evidence to support such partitions of the data. However, such partitions do not lead to a parsimonious set of well-separated clusters.

For $R_{\text{ATA-198}}$ there was no clear validity index consensus. Some indices favored $c=6$ or $c=8$ clusters, which is not completely consistent with a visual inspection of the groups highlighted by the re-ordered relationships. Other indices suggested that $c=12$ or $c=14$ clusters best describe the data, which better aligns with our results. Our previous studies with this data indicate that there are viable explanations for each of these cluster counts.

## 4. CONCLUSIONS

In this paper, we have proposed a deterministic-annealing-based approach to relational clustering. Our approach is based upon producing a type of partition matrix, known as an accumulation matrix, that quantizes the original relational matrix. We rely on an information-theoretic criterion, the value of information, to specify a computationally feasible procedure for finding globally optimal accumulation matrices. The value of information trades off against the amount of information against the quantization fit of the accumulation matrix to the original relational data.

Ranges of information amounts lead to different number of clusters. The information amount also dictates the fuzziness of the partitions. Both of these properties are data-dependent: the best values for one dataset may not work for another. We hence provided a heuristic for choosing the 'right' amount of information, and hence a parsimonious partition, in a data-driven fashion; no parameters need to be set by investigators when using our clustering approach. This heuristic performed well for the complicated datasets that we considered.

---

[1]https://www.dropbox.com/s/w2qhu63fvlz1otc/Sledge-ICASSP-2017-2col-appendix.pdf?dl=0